\documentclass[a4paper,fleqn]{cas-dc}

\usepackage[numbers]{natbib}
\usepackage{graphicx, subfig}
\usepackage{algorithm}
\usepackage{algorithmic}
\usepackage{amsmath}
\usepackage{mathtools}
\usepackage{amsthm}
\newtheorem{example}{Example}
\usepackage{color}
\usepackage{multirow}
\usepackage{makecell}
\newcommand\inter[2]{\mathcal{I}_{\mathcal{#1}}(#2)}

\def\tsc#1{\csdef{#1}{\textsc{\lowercase{#1}}\xspace}}
\tsc{WGM}
\tsc{QE}

\begin{document}
\let\WriteBookmarks\relax
\def\floatpagepagefraction{1}
\def\textpagefraction{.001}

\shorttitle{}    

\shortauthors{Chen, J., Ren, J., Ding, W. et al.}

\title [mode = title]{Conflict Detection for Temporal Knowledge Graphs:A Fast Constraint Mining Algorithm and New Benchmarks}  



\author[]{Jianhao Chen}[orcid=0009-0006-4363-0549]

\ead{jh_chen@smail.nju.edu.cn}

\credit{Conceptualization, Methodology, Software, Writing – original draft}

\affiliation{organization={State Key Laboratory for Novel Software Technology, Nanjing University},
            city=Nanjing,
            country=China}

\author[]{Junyang Ren}[]
\ead{jyren@smail.nju.edu.cn}      
\credit{Data Curation, Resources}

\author[]{Wentao Ding}[]
\ead{wtding@smail.nju.edu.cn}
\credit{Conceptualization, Writing – original draft}

\author[]{Haoyuan Ouyang}[]
\ead{hyouyang@smail.nju.edu.cn}
\credit{Validation}

\author[]{Wei Hu}[orcid=0000-0003-3635-6335]
\ead{whu@nju.edu.cn}
\credit{Writing – review \& editing}

\author[]{Yuzhong Qu}[orcid=0000-0003-2777-8149]
\ead{yzqu@nju.edu.cn}
\credit{Supervision}
\fnmark[*]


\cortext[0]{* Corresponding author}


\begin{abstract}
Temporal facts, which are used to describe events that occur during specific time periods, have become a topic of increased interest in the field of knowledge graph (KG) research. In terms of quality management, the introduction of time restrictions brings new challenges to maintaining the temporal consistency of KGs. Previous studies rely on manually enumerated temporal constraints to detect conflicts, which are labor-intensive and may have granularity issues. To address this problem, we start from the common pattern of temporal facts and propose a pattern-based temporal constraint mining method, PaTeCon. Unlike previous studies, PaTeCon uses graph patterns and statistical information relevant to the given KG to automatically generate temporal constraints, without the need for human experts. In this paper, we illustrate how this method can be optimized to achieve significant speed improvement. We also annotate Wikidata and Freebase to build two new benchmarks for conflict detection. Extensive experiments demonstrate that our pattern-based automatic constraint mining approach is highly effective in generating valuable temporal constraints.
\end{abstract}




\maketitle

\section{Introduction}
\label{Introduction}
Knowledge graphs (KGs) are representations of real-world facts that plays an important role in many AI applications, including information retrieval, natural language question answering, and recommender systems. However, classical KGs provide only a static view of the real world through RDF triples in the form of subject-property-object ($(S,P,O)$ in short) and cannot meet the specific time demands of downstream applications. For instance, classical KGs may contain a fact such as (\textsf{Lionel Messi, play for, FC Barcelona}) to record Messi's career in FC Barcelona, but this fact alone cannot answer questions like ``When did Lionel Messi play for FC Barcelona?". The growing needs for temporal facts modeling and acquisition in KGs have received increasing attention in recent years. A temporal fact can be formalized as a quadruple $(S,P,O,T)$, where the time interval $T$ indicates the period during which the subject-property-object triple $(S,P,O)$ holds.

In practical applications, popular KGs such as Freebase~\cite{FREEBASEsigmod08} and Wikidata~\cite{WIKIDATAacm14} have incorporated numerous temporal facts through specially designed representation structures. However, despite their widespread use, practical KGs still suffer from quality issues and incompleteness. Example~\ref{eg:conflict} provides a concrete instance of a real conflict in Wikidata, underscoring the need for better mechanisms to identify and address such conflicts.
\begin{example}\label{eg:conflict}
Wikidata, one of the most widely used KG, contains two temporal facts\footnote{\url{https://www.wikidata.org/wiki/Q3067041}} as follows:
\begin{align*}
&\left(\mathsf{Q3067041},\mathsf{member\dots}, \mathsf{Q15301637},[1988,1992]\right),\\
&\left(\mathsf{Q3067041},\mathsf{member\dots}, \mathsf{Q20738291},[1989,1990]\right).
\end{align*} 
$\mathsf{Q3067041}$ is the Wikidata ID for ``Farès Bousdira'', a football player. $\mathsf{Q15301637}$ and $\mathsf{Q20738291}$ are the Wikidata IDs of ``CS Saint-Denis'' and ``AS Possession'', respectively, two association football team. $\mathsf{member\dots}$ is the abbreviation for wikidata property $\mathsf{member\_of\_sports\_team}$.
It is obvious that they cannot both hold because ``\textit{A player cannot play for two association football teams at the same time}''.
\end{example}

Given the current state of practical KGs, there is a pressing need for acquisition and quality management of temporal knowledge. The inclusion of temporal restrictions has introduced a new challenge to KG quality management: the potential for inconsistencies in the time dimension. Moreover, in recent years, deep learning-based methods have gained significant popularity for knowledge acquisition. However, since these approaches cannot offer theoretical guarantees for the consistency of the acquired knowledge, it may be necessary to conduct explicit quality checks on the acquired knowledge.

To ensure the temporal consistency of KGs, it is crucial to detect conflicting temporal facts. Previous examples have highlighted that people often identify temporal conflicts by recognizing violations of particular constraints. For instance, \citet{AAAI17} provides an example of such a rule, as demonstrated in Example~\ref{eg:raw_rule_sports_teams}. By identifying and listing down the temporal constraints that apply to a given KG, we can readily identify all possible conflicts that breach these constraints.

\begin{example}
\label{eg:raw_rule_sports_teams}
Rule for representing the constraint ``\textit{One cannot be a member of two sports teams at the same time}'':
\begin{align}
\label{f:eq1}
\begin{split}
\mathsf{disjoint}\left(t_1,t_2\right)  \coloneq &
\left(x,\mathsf{member\_of\_sports\_team},y,t_1\right), \\
& \left(x,\mathsf{member\_of\_sports\_team},z,t_2\right), \\
& \ y\neq z . 
\end{split}
\end{align}
\end{example}

Previous research has relied on human experts to list temporal constraints. While experienced experts can produce high-quality constraints, the manual process of enumerating temporal constraints is time-consuming. Additionally, human experts may encounter difficulties in determining the appropriate level of detail for these temporal constraints. For example, the constraint presented in Formula~(\ref{f:eq1}) by \citet{AAAI17} fails to consider that the range of the $\mathsf{member\_of\_sports\_team}$ property is actually $\mathsf{sports\_organization}$, which encompasses both national teams and clubs. As a result, it may mistakenly identify "an athlete who is a member of a club while being a member of the national team" as a temporal conflict. Identifying these exceptions manually would be tedious, if not impossible. Furthermore, manual constraint engineering is limited in its applicability. The constraints identified manually cannot be automatically updated when the KG is updated, nor can they be easily transferred to other KGs. Therefore, there is a pressing need for an effective and general technique for automatically mining temporal constraints.

The goal of this paper is to mine temporal constraints from KGs. As demonstrated in Formula~(\ref{f:eq1}), previous research indicates that logical rules can be utilized to represent temporal constraints. Our findings further indicate that these rules can be decomposed into graph patterns consisting of KG facts and temporal statements about them. For instance, the rule's body part in Formula~(\ref{f:eq1}) matches two facts having the same subject and property, but distinct objects. Meanwhile, its head part is a temporal statement that asserts the non-overlapping nature of the involved time intervals. Identifying temporal conflicts entails identifying all subgraphs that match these patterns.

\begin{figure*}
    \centering
    \subfloat[]{\includegraphics[scale=0.8]{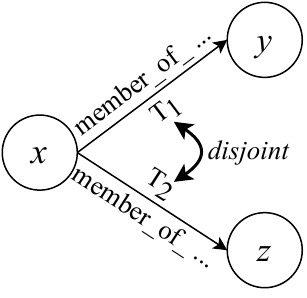}}
    \hspace{25mm}
    \subfloat[]{\includegraphics[scale=0.8]{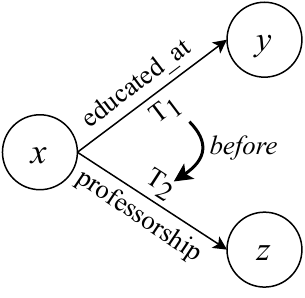}}
    \caption{Examples of the graph representation of temporal constraints. Note that different variables (e.g., $y$ and $z$) should correspond to different things if not specified.}
    \label{fig:2tc}
\end{figure*}

\begin{figure}
    \centering
    \includegraphics[scale=0.7]{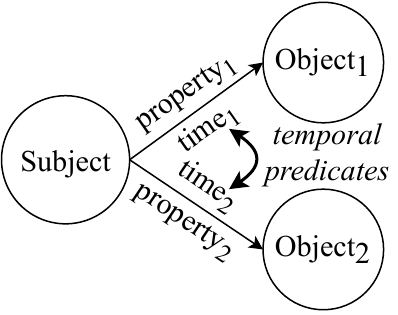}
    \caption{The common structural pattern of the temporal constraints illustrated in Figure~\ref{fig:2tc}.}
    \label{fig:cSP}
\end{figure}

This approach directs our attention to the shared patterns among different constraints. For example, the temporal constraint on athletes and another constraint ``one must be educated somewhere before his professorship" can be modeled as shown in Figure~\ref{fig:2tc}, and they share a common structural pattern depicted in Figure~\ref{fig:cSP}.

Our previous work \textbf{PaTeCon}~\cite{chen2023PaTeCon} has shown how to utilize patterns to mine temporal constraints, which eliminates the need for human intervention to obtain specific temporal constraints and can be applied to any temporal KG without restriction. PaTeCon leverages statistical information from instances of predefined structural patterns to extract potential constraints. In particular, it enhances coarse constraints by incorporating class restrictions. The experimental results show that pattern-based automatic constraint mining is powerful in generating valuable temporal constraints. Besides, PaTeCon could run on KGs with more than 10M facts within hours, which is an acceptable running time.

In our previous work, two large-scale datasets (WD27M and FB37M) were constructed for constraint mining. However, both datasets are unannotated. This means that we cannot automatically evaluate the conflict detection task on these two datasets. As far as we know, past conflict detection work~\cite{AAAI17} on temporal KGs and error detection work~\cite{KGTtm,PGE,CAGED,KAEL} on KGs all generate wrong facts by replacing entities or properties. We argue that the wrong facts obtained by replacement cannot simulate the true distribution of wrong facts in KGs. We thus annotate wrong facts in real-world KGs for the first time.

This paper significantly extends PaTeCon in four aspects: (1) We design a pruning strategy to substantially speed up the constraint mining algorithm and refer to this new version as PaTeCon+\footnote{Source code and documentation can be accessed at \url{https://github.com/JianhaoChen-nju/PaTeCon}.}. (2) We design experiments to illustrate the difference between constraints under entity-level confidence and fact-level confidence measures. (3) We semi-automatically annotate WD27M and FB37M and build wrong fact sets WD-411 and FB-128, respectively. We perform an automatic evaluation on these two benchmarks. (4) We extend the application scenario, which utilizes constraints mined from existing KGs to detect conflicts between new facts and original KGs.

In summary, our main contributions in this paper are outlined as follows:
\begin{enumerate}
\item We provide a more in-depth analysis and experimental evaluation of the design of the entity-level confidence in PaTeCon.
\item We design a two-stage pruning strategy, enabling our method to quickly mine KGs with more than 10M facts.
\item We develop two new benchmarks for conflict detection through semi-automatic construction.
\item We verify the feasibility of using constraints mined in KGs to enhance the quality of newly added facts.
\end{enumerate}

The remainder of this paper is organized as follows. Section~\ref{sec:pre} introduces the preliminaries, including the explanation of temporal facts, time intervals, and temporal constraints along with their structural patterns. Section~\ref{sec:em} presents a comprehensive description of evaluation metrics. In Section~\ref{sec:PaTeCon}, we present the framework and our
implementation of PaTeCon. In Section~\ref{sec:speed}, the PaTeCon+ pruning strategy is elaborated in detail, which aims to enhance the speed of the method. Section~\ref{sec:exp} showcases the evaluation of PaTeCon+ with two new large-scale datasets. Section~\ref{sec:work} discusses the related work, and finally, Section~\ref{sec:con} concludes the paper and discusses future work.

\section{Preliminaries}
\label{sec:pre}

\subsection{Knowledge Graph with Temporal Facts}
In this paper, the definition of \textit{knowledge graph (KG)} refers to knowledge bases that are represented using the RDF data model. A classical RDF KG expresses knowledge through a set $\mathcal{R}$ of IRI resources (resources with identifiers) and a set $\mathcal{L}$ of literals (string data), with the conceptual terms of the KG being modeled by classes $\mathcal{C} \subseteq \mathcal{I}$ and properties $\mathcal{P} \subseteq \mathcal{I}$ of the IRI resources.
An atomic fact in the RDF KG is a triplet $F=(S, P, O) \in \mathcal{R}\times\mathcal{P}\times (\mathcal{R} \cup \mathcal{L})$. Additionally, RDF allows for the representation of complex facts (e.g., those with time restrictions) via anonymous resources, a.k.a. RDF blank nodes.
To express the theoretical model for temporal constraints mining, the term \textit{temporal facts} is used to denote complex facts with time restrictions. A temporal fact is a quadruple $TF=(S, P, O, T) \in \mathcal{R}\times\mathcal{P}\times (\mathcal{R} \cup \mathcal{L})\times (\mathcal{T} \times \mathcal{T})$, where $\mathcal{T}$ denotes the time domain and $T=(t.s, t.e)$ represents a time interval defined by its start and end times.

\subsection{Time Intervals and Their Algebra}
An ordered pair of time values, $t.s$ and $t.e$, denotes the start and end points of a time interval $T$. By comparing the endpoints of two time intervals, we can determine their relation to interval algebras such as \citet{ALLEN83}. To generate temporal constraints, we use five temporal predicates: $\mathsf{start}$, $\mathsf{finish}$, $\mathsf{before}$, $\mathsf{disjoint}$, and $\mathsf{include}$. Section~\ref{sec:ut} presents the calculation details of these predicates, which are chosen according to practical demands.

\begin{figure*}[t]
    \centering
	\subfloat[]{\includegraphics[scale=0.75]{figures/pattern_structure_2.pdf}}
	\hspace{25mm} 
	\subfloat[]{\includegraphics[scale=0.75]{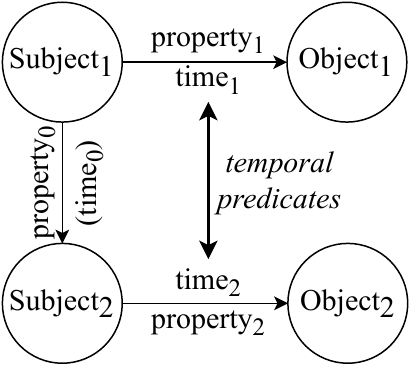}}
    \caption{Structural patterns. Noting that the properties could be reversed.}
    \label{fig:structural patterns}
\end{figure*}

\subsection{Temporal Constraints}
A temporal constraint is a logical assertion of the time dimension. We formalize a temporal constraint $tc$ in the form of $tc = H\coloneq B$, where the head $H$ is an atomic predicate, and the body $B$ is a conjunction of atoms with optional variables.
In this paper, we consider three kinds of temporal constraints, which cover most constraint types in previous studies~\cite{AKBC14,AAAI17,ETCwise18}.

\textbf{Temporal Disjointness} For some temporal properties, the time intervals of two facts with one common subject should be non-overlapping. Furthermore, the disjointness between a pair of time intervals with different properties is trivial, so we only consider the case of the same property in this paper:
\begin{align}
\mathsf{disjoint}(t_1,t_2) \coloneq &
\left(x,\mathsf{p},y,t_1\right), \left(x,\mathsf{p},z,t_2\right), y\neq z. 
\end{align}
Example~\ref{eg:raw_rule_sports_teams} has provided a concrete instance of a temporal disjointness constraint.

\textbf{Temporal Ordering} Temporal ordering depicts the temporal relationships ($\mathsf{start}$, $\mathsf{finish}$, $\mathsf{before}$,  and $\mathsf{include}$) between two experience of one or more subjects.
\begin{align}
\mathsf{order}\left(t_1,t_2\right) \coloneq &
\left(x,\mathsf{p_1},y,t_1\right),
\left(x,\mathsf{p_2},z,t_2\right). \\
\mathsf{order}\left(t_1,t_2\right) \coloneq &
\left(x,\mathsf{p_1},z,t_1\right),
\left(x,\mathsf{p_0},y\right),
\left(y,\mathsf{p_2},w,t_2\right).
\end{align}
Example~\ref{eg:patternB} provides a concrete instance of a temporal ordering constraint.
\begin{example}\label{eg:patternB}
Rule for representing the constraint ``A person cannot be killed by someone who is already dead''. 
\begin{align}
\begin{split}
\mathsf{before}\left(t_1,t_2\right) \coloneq  & 
\left(x,\mathsf{place\_of\_death},z,t_1\right),\\
&\left(x,\mathsf{killed\_by},y\right),\\
&\left(y,\mathsf{place\_of\_death},w,t_2\right).
\end{split}
\end{align}
\end{example}

\textbf{Mutual Exclusion} Mutual exclusion defines a special set of facts which are in conflict with each other regardless of time. Specifically, it means that the property $\mathsf{p}$ of a subject $x$ can only have one object:
\begin{align}
\mathsf{false} & \coloneq 
\left(x,\mathsf{p},y,t_1\right), \left(x,\mathsf{p},z,t_2\right), y\neq z  \label{f:eq2}.
\end{align}
Example~\ref{eg:me} provides a concrete instance of a mutual exclusion constraint.
\begin{example}\label{eg:me}
``One can only die once'', thus one person should have only one death place. 
\begin{align}
\begin{split}
\mathsf{false}   \coloneq &
\left(x,\mathsf{place\_of\_death},y,t_1\right),\\
&\left(x,\mathsf{place\_of\_death},z,t_2\right),\\
&\ y\neq z. \label{f:eq3}
\end{split}
\end{align}
\end{example}

\subsection{Structural Patterns}\label{sec:sp}
The structural pattern perspective shows that a temporal constraint can be divided into two parts: (1) the triple/quadruple patterns for facts in KGs and (2) the constraint predicate over the facts.  
In practical applications, we use quadruples to match temporal facts in KGs. For example, the quadruple 
\begin{align*}
\left(\mathsf{Lionel\_Messi},\mathsf{member\_of\_sports\_team},x,T\right)
\end{align*}
can match all of Lionel Messi's experiences as an athlete. We use $\mathcal{I}_{\mathcal{G}}(tc)$ to denote the subgraphs of $\mathcal{G}$ satisfying the body (i.e., the graph pattern) of $tc$. 

In this paper, we model temporal constraints via their common patterns, i.e., the structural patterns. These patterns depict the graph structure of facts in particular constraints and indicate the resources involved in the temporal predicates. Figure~\ref{fig:structural patterns} illustrates two fundamental structural patterns employed by our method.

By utilizing these structural patterns, we can capture temporal relationships between two subjects or two experiences of one subject. As discussed in the introduction, structural patterns offer the flexibility to model various types of constraints by modifying the filling-in predicates.

\begin{figure}
    \centering
	\subfloat[]{{\includegraphics[scale=0.85]{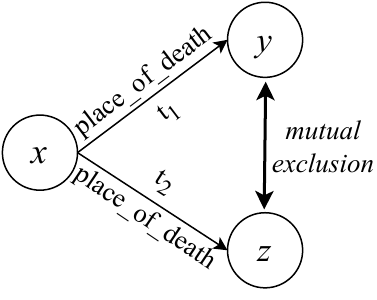}}}
	\hspace{20mm} 
	\subfloat[]{\includegraphics[scale=0.85]{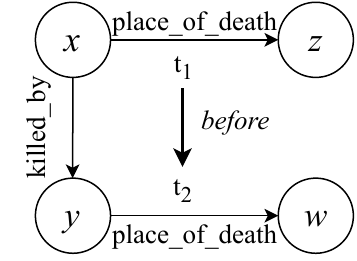}}
    \caption{Graph representations of Example~\ref{eg:me} and Example~\ref{eg:patternB}.}
    \label{fig:mutex}
\end{figure}

Structural pattern (a), for instance, can effectively represent most of the temporal relationships including temporal disjointness and ordering. Moreover, it can also represent non-interval relationships such as mutual exclusion, which denotes the uniqueness of certain events. Figure~\ref{fig:mutex}(a) shows that such constraints like Example~\ref{eg:me} can be modeled via structural pattern (a) by taking $\mathsf{false}$ as a special atom in the predicate slot.

Structural pattern (b) further considers the assertions about multiple subjects. It assumes that two temporally-related subjects must be connected in the KG. An example of expressing such constraints using structural pattern (b) is shown in Example~\ref{eg:patternB}. Figure~\ref{fig:mutex}(b) shows that this constraint can be modeled via structural pattern (b) by filling $\mathsf{killed\_by}$ into $property_0$ slot and $\mathsf{place\_of\_death}$ into slots $property_1$ and $property_2$.

However, to keep the search manageable, we do not extract the disjointness relationships between different subjects in practical implementation. For more intricate constraints, such as those involving three subjects, we can create them by combining the constraints of simpler patterns. Therefore, we do not need to create structural patterns specifically for these cases.

\section{Evaluation Metrics}
\label{sec:em}

Support and confidence are the most commonly used metrics to measure the quality of rules. In this section, we describe the measures for judging rule quality in detail. We first describe the challenges of defining such metrics in our setting, and how to calculate logical values for temporal predicates over uncertain time intervals. Then, we discuss the most common way to measure the quality of rules, which we refer to as fact-level support and confidence. At last, we introduce our own metric: entity-level support and confidence.

\subsection{Challenges}
To compute the support and confidence of a temporal constraint, we have to calculate the logical value of the temporal predicate. 
Datasets in previous work~\cite{AAAI17,Jiangcoling16} on temporal KGs often express time as a unified granularity (for example, day), and do not select the facts in which start time or end time is missing. However, the granularity of time in real KGs is often not uniform. Practically speaking, KGs cannot always provide precise occurrence periods for events. For example, Wikidata records that the $\mathsf{start\_time}$ and $\mathsf{end\_time}$ of Barack Obama's employment as a Sidley Austin employee are both 1991, which does not mean that the $\mathsf{start\_time}$ and $\mathsf{end\_time}$ are precisely equal, nor does it mean that Obama held that job throughout the whole 1991. In addition, there are often cases where the start time or end time is missing. For example, the end time may be missing because the event is not recorded in detail or the event has not ended yet. Therefore, the first major challenge is to calculate the logical values of temporal predicates over uncertain time intervals.

Temporal constraint mining is to discover which temporal fact is anomalous in the KG. Therefore, if we consider a temporal constraint to be of better quality, temporal facts that do not satisfy the constraint are more likely to be outliers. 
The measurement of the quality of a temporal constraint $tc$ on a given KG $\mathcal{G}$ can be performed on $\inter{G}{tc}$, which denotes the subgraphs of $\mathcal{G}$ satisfying the body (i.e., the graph pattern) of $tc$. Thus, the second challenge is what kind of subgraph can be considered as a support, and the support and confidence calculated from it must reflect regularity as closely as possible. We introduce two ways to compute support and confidence: fact-level support and confidence, and our proposed entity-level support and confidence. We now present these metrics in detail.

\begin{table*}[htb]
    \centering
    \caption{Calculation of temporal predicates over uncertain temporal intervals.}
    
\begin{tabular*}{\linewidth}{LLLL}
    \hline\hline
    Predicates & Positive Condition & Negative Condition & Unknown\\
    \hline
    \multirow{3}*{$\mathsf{start}$} & \multirow{3}*{$T_1.s = T_2.s$} & $(T_1.s \ne T_2.s)$ & \multirow{3}*{otherwise}\\
        &   & $\vee (T_1.s > T_2.e)$ & \\
        &   & $\vee (T_1.e < T_2.s)$ & \\
    \hline
    \multirow{3}*{$\mathsf{finish}$} & \multirow{3}*{$T_1.e = T_2.e$} & $(T_1.e \ne T_2.e)$ & \multirow{3}*{otherwise}\\
        &   & $\vee (T_1.e < T_2.s)$ &  \\
        &   & $\vee (T_1.s > T_2.e)$ &  \\
    \hline
    \multirow{4}*{$\mathsf{before}$} & \multirow{4}*{\makecell[l]{$(T_1.e<T_2.s)$\\$\vee (T_1.e=T_2.s \wedge T_1 \ne T_2)$}} & $(T_1.e > T_2.s)$ & \multirow{4}*{otherwise}\\
        &   & $\vee (T_1.e > T_2.e)$ &  \\
        &   & $\vee (T_1.s > T_2.s)$ &  \\
        &   & $\vee (T_1.s > T_2.e)$ &  \\
    \hline
    \multirow{2}*{$\mathsf{disjoint}$} & $(T_1\ before\ T_2)$ & $(T_1.e>T_2.s \wedge (T_1.e \leq T_2.e \vee T_1.s \leq T_2.s))$ & \multirow{2}*{otherwise}\\
        & $\vee (T_2\ before\ T_1)$ & $\vee (T_1.s<T_2.e \wedge (T_1.e \geq T_2.e \vee T_1.s \geq T_2.s))$ &  \\
    \hline
    \multirow{4}*{$\mathsf{include}$} & \multirow{4}*{\makecell[l]{$(T_1.s \leq T_2.s)$\\$\wedge (T_2.e \leq T_1.e)$}} & $(T_1.s>T_2.s)$ & \multirow{4}*{otherwise}\\
        &   & $\vee (T_1.s > T_2.e)$ &  \\
        &   & $\vee (T_1.e < T_2.e)$ &  \\
        &   & $\vee (T_1.e < T_2.s)$ &  \\
    \hline\hline
\end{tabular*}
    \label{tab:ctp}
\end{table*}

\begin{table}[htb]
    \centering
    \caption{Calculation on (possibly absent) time values of different granularity.}
    \begin{tabular*}{\linewidth}{LLLL}
        \hline\hline
        $t_1$ & $t_2$ & $t_1 < t_2$ & $t_1 = t_2$  \\
        \hline
        2021-12 & 2022 & Positive & Negative \\ 
        2022-01 & 2022 & Unknown & Unknown \\ 
        - & 2022 & Unknown & Unknown \\
        \hline\hline
    \end{tabular*}
    \label{tab:comp}
\end{table}
\subsection{Temporal Predicates over Uncertain Time Intervals}
\label{sec:ut}

In this subsection, we explain in detail the process of calculating logical values for temporal predicates. The head predicates' logical values are calculated during time intervals represented as $T=(t.s, t.e)$. The starting time is denoted by $t.s$ and the ending time by $t.e$. 
To deal with the absence and granularity of time values, we set the logical values of time predicates in $tc$ to positive, negative, or unknown.

Table~\ref{tab:ctp} shows the positive, negative, and unknown conditions of the predicates. If the granularity is not sufficiently fine or some time values are missing, the calculation result would be marked as \textit{unknown} instead of \textit{negative}. Table~\ref{tab:comp} provides an illustration of this.

\subsection{Fact-level Support and Confidence}
\begin{figure}[htb]
    \centering
    \includegraphics[scale=0.4]{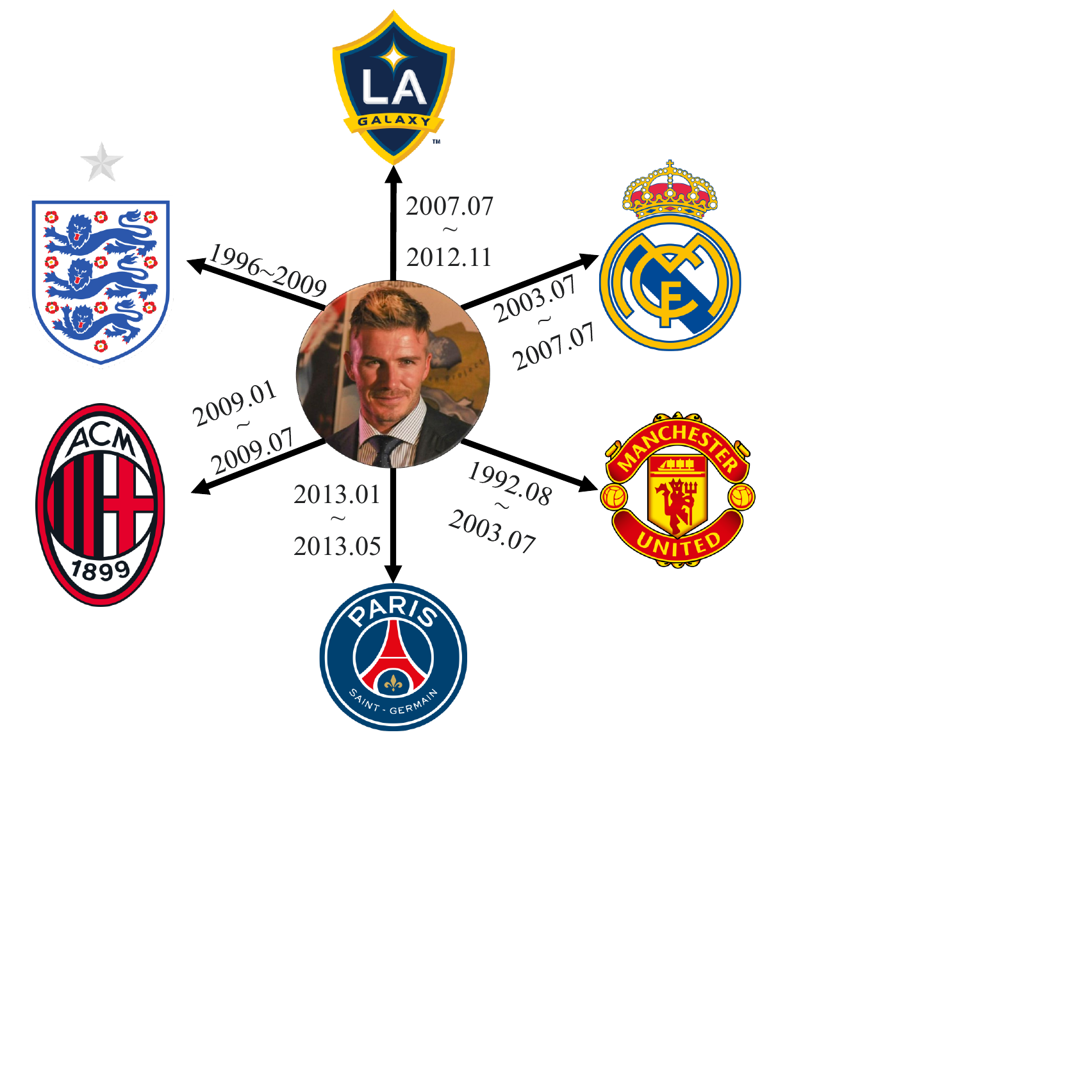}
    \caption{Part of David Beckham's career recorded at Wikidata. All predicates omitted by the arrows on the graph are \textsf{member\_of\_sports\_team}.}
    \label{fig:beckham}
\end{figure}

Under the fact-level metric, an instantiation is a subgraph $F$ (i.e., a collection of facts) that can match the body of the constraint. For instance, in Figure~\ref{fig:beckham} (\textsf{Beckham, member of sports team, Manchester United F.C., [1992.08, 2003.07]}), (\textsf{Beckham, member of sports team, Real Madrid CF, [2003.07, 2007.11]}) form an instantiation (we refer to this instantiation as $ins_1$ below) of the temporal constraint in Formula~(\ref{f:eq1}): 
\begin{align*}
\mathsf{disjoint}\left(t_1,t_2\right) \coloneq &
\left(x,\mathsf{member\_of\_sports\_team},y,t_1\right), \notag\\
& \left(x,\mathsf{member\_of\_sports\_team},z,t_2\right), \notag\\
& \ y\neq z.  
\end{align*}
We refer to it as $tc_1$ below.

The fact-level support counts the number of positive conditions of instantiations:
\begin{align}
support_{fact-level} = \#F_{positive}. \label{f:std supp}
\end{align}
For instantiation $ins_1$, logical value of $H$ \textsf{disjoint}([1992.08, 2003.07], [2003.07, 2007.11]) is positive. So this pair of facts and their positive value of temporal predicate form a support for constraint $tc_1$.
Under this definition, we can count $C(6, 2)=15$ instantiations in Figure~\ref{fig:beckham}, and 10 of those 15 instantiations are positive. 4 of those 15 instantiations are negative. Specifically, note that the logical value of \textsf{disjoint}([1996, 2009], [2009.01, 2009.07]) is unknown, so (\textsf{Beckham, member of sports team, England national association football team, [1996, 2009]}), (\textsf{Beckham, member of sports team, A.C. Milan, [2009.01, 2009.07]}) cannot form a support. We only use positive and negative instantiations to calculate support and confidence scores.
Thus, the fact-level confidence of a $tc$ is defined as follows:
\begin{align}
confidence_{fact-level} =\frac{\#F_{positive}}{\#F_{positive}+\#F_{negative}}. \label{f:std confidence}
\end{align}
The fact-level confidence of $tc_1$ calculated from Figure~\ref{fig:beckham} is $\frac{5}{7}$.

However, here is an extreme example: The fact-level confidence of constraint ``\textit{Two or more people cannot play for a team at the same time}'' is up to 0.94 in Wikidata. This may sound absurd but let us explain how this constraint scores high under fact-level confidence calculations. Imagine a veteran team that has many players from its inception to the present, but does not have many players at the same time. If two players are chosen from all the players who have played for the team, the probability that their time on the team is disjoint would be 0.94.

\subsection{Entity-level Support and Confidence}
\label{sec:entitylev}
In PaTeCon, we regard temporal facts as the activity of entities.  Temporal constraints essentially reflect the regularity of entity activities. Therefore, we propose an entity-level instead of a fact-level confidence calculation method. We regard the entire graph in Figure~\ref{fig:beckham} as an instantiation, which describes the career of an athlete. 
Specifically, we have
\begin{align}
entities = \bigcup_{F \in \inter{G}{tc}} subject(F),
\end{align}
where $F$ denotes the subgraph matched by the body of $tc$, $subject(F)$ denotes the entity in $F$ that are matched by the $subject$ and $subject_1$ slots in structural pattern (a) and (b), respectively.
Under this definition, there is only one entity in Figure~\ref{fig:beckham}, which is \textsf{Beckham}.

When classifying entities, an important challenge is dealing with the uncertainty of KGs in the time dimension. To address this issue, we divide entities into the positives $entities_{pos}$, the negatives $entities_{neg}$, and the unknowns $entities_{unk}$ based on the logical value of the head predicate in $tc$. An entity is positive if all of the logical values for the matched subgraphs are positive, and negative if any of the logical values for the matched subgraphs are negative. Besides, an entity is unknown if none of the logical values for the matched subgraphs are negative and at least one of them is unknown. We regard the positives as support, i.e.,
\begin{align}
support_{entity-level} = \#entities_{pos},\label{f:entity-support}
\end{align}
and use the positives and negatives to calculate confidence scores, while the unknowns are disregarded.
Formally, the \textit{entity-level confidence} is determined by the following definition:
\begin{align}
confidence_{entity-level} = \frac{\#entities_{pos}}{\#entities_{pos} + \#entities_{neg}},\label{f:conf}
\end{align}
where $entities_{pos}$ and $entities_{neg}$ are the subsets of subject entities in $\inter{G}{tc}$.
Under this definition, the entity-level confidence of $tc_1$ calculated from Figure~\ref{fig:beckham} is $\frac{0}{1}$.

Next, we theoretically explain the difference between entity-level confidence and fact-level confidence. Experiments in Section~\ref{sec:confidencevs} show the performance differences of the temporal constraints that they mine. It can be seen from the definition that we aggregate subgraphs with common entities into a larger subgraph, which inevitably leads to a reduction in the number of supports and instantiations. Since only entities that match all logical values of the subgraph are positives, the condition for an entity to be positive is very strict. Therefore, the score calculated by the entity-level confidence is often lower than the fact-level confidence (e.g., in our case illustrated above 0 is less than 5/7). Temporal constraints that still achieve high scores under entity-level confidence metrics can better reflect the real regularity of entity activities.

In addition, when calculating the entity-level confidence, as long as the logical value of the temporal predicate of the subgraph is calculated to be negative, there is no need to calculate the remaining subgraphs. This can well reduce the overhead of confidence calculation. 

\section{PaTeCon}
\label{sec:PaTeCon}

The framework of our method, PaTeCon, is outlined in Algorithm~\ref{alg:algorithm}. Initially, the method extracts frequent candidate constraints using pre-defined structural patterns $\mathcal{SP}$. Subsequently, it calculates the confidence scores of these candidates in order to refine them dynamically and obtain the final set of temporal constraints.
Subsequent subsections provide further information about the crucial aspects of candidate mining, confidence calculation, and constraint refinement.

\begin{algorithm}[ht]
\caption{Temporal Constraint Mining}
\label{alg:algorithm}
\textbf{Input}:the KG $\mathcal{G}$, structural patterns $\mathcal{SP}$
and temporal predicates $\mathcal{TP}$\\
\textbf{Output}:the constraints $\mathcal{TC}$
\begin{algorithmic}[1]
\STATE $Candidates = \emptyset$ \hfill\textit{// Mining candidate constraints.}
\STATE $\mathcal{GP} = \mathrm{InstantiateToGraphPatterns}(\mathcal{G},\mathcal{SP})$
\FOR{$gp \in \mathcal{GP}$}
    \STATE $sg = \mathrm{MatchedSubGraphs}(\mathcal{G},gp)$
    \FOR{$tp \in \mathcal{TP}$}
        \STATE $tc = \mathrm{GenerateConstraint}(tp, gp)$
        \STATE $s = \mathrm{ComputeSupport}(sg, tc)$
        \IF{$s < \theta_{freq}$}
            \STATE \textbf{continue}
        \ENDIF
        \STATE $Candidates = Candidates \cup \{tc\}$
    \ENDFOR
\ENDFOR
\STATE $\mathcal{TC}=\emptyset$ \hfill\textit{// Obtaining high-quality constraints.}
\FOR{$tc \in Candidates$}
    \STATE $gp = tc.\mathrm{graph\_pattern}$ 
    \STATE $sg = \mathrm{MatchedSubGraphs}(\mathcal{G},gp)$
    \STATE $c = \mathrm{ComputeConfidence}(sg, tc)$
    \IF{$c > \theta_{c1}$}
        \STATE $\mathcal{TC}=\mathcal{TC} \cup \{tc\}$
    \ELSIF{$c > \theta_{c2}$}
        \STATE $\mathcal{RC} = \mathrm{RefineConstraint}(tc)$
        \FOR{$rc \in \mathcal{RC}$}
            \IF{$\mathrm{ComputeSupport}(sg, rc) > \theta_{freq} \wedge \mathrm{ComputeConfidence}(sg, rc) > \theta_{c1}$}
                \STATE $\mathcal{TC}=\mathcal{TC} \cup \{tc\}$
            \ENDIF
        \ENDFOR
    \ENDIF
\ENDFOR
\STATE \textbf{return} $\mathcal{TC}$
\end{algorithmic}
\end{algorithm}

\subsection{Mining Candidate Constraints}
The first 10 lines of the algorithm cover the process of identifying frequent candidates. To achieve this, PaTeCon starts by converting pre-defined structural patterns $\mathcal{SP}$ into graph patterns $\mathcal{GP}$ that have been observed in the KG. This conversion involves filling in the \textit{property} slots in the $SP$s based on the connection structure of the facts in $\mathcal{G}$. Once the graph patterns have been matched, temporal predicates are added to them to generate all possible temporal constraints. Only the constraints with support $s$ that meet or exceed the threshold $\theta_{freq}$ would be selected as candidates. The instantiation process involves selecting all entities in $\mathcal{G}$ as possible subjects in $SP$s and then searching the neighboring subgraphs to obtain the relevant properties.

\subsection{Refining Constraints}
As mentioned earlier in the introduction, constraints that only have property restrictions may not be precise enough for detecting conflicts. In cases where the confidence $c$ of a constraint $tc$ is below the quality threshold $\theta_{c_1}$ but above a more lenient threshold $\theta_{c_2}$, we make an effort to narrow down its domain by class within $\mathcal{G}$. Specifically, we try to identify the different combinations of classes for the related entities in the matched subgraphs $\mathcal{SG}$. Any refined constraints that meet the criteria of having a support of at least $\theta_{freq}$ and a confidence of at least $\theta_{c_1}$ would be included in the final results.

\section{Speed Improvement: PaTeCon+}
\label{sec:speed}

Based on the original PaTeCon framework~\cite{chen2023PaTeCon}, we extend it in this paper with a two-stage pruning strategy to speed up the constraint mining phase. In the following, we describe the pruning strategy in detail and refer to this new version of PaTeCon as PaTeCon+. In Section~\ref{sec:assumpt} we discuss the underlying assumption and in Section~\ref{sec:pruning} we explain in detail how it is used within PaTeCon+.

\subsection{Underlying Assumption}
\label{sec:assumpt}

The overhead of our algorithm is mainly concentrated in two stages: matching subgraphs and computing confidence scores. To compute confidence scores for constraints, we need to match all subgraphs that share a common subject, concatenate with temporal predicates to generate possible constraints, and compute support and confidence.
At the same time, most of the generated constraints do not meet our final confidence threshold. This means that the system would spend a lot of time on matching subgraphs and computing the confidence scores for those generated constraints having no value. An example of such generated constraints is as follows:
\begin{align}
\label{worthlessconstraint}
\begin{split}
\mathsf{before}\left(t_1,t_2\right)  \coloneq  &
\left(x,\mathsf{residence},y,t_1\right), \\
&\left(x,\mathsf{award\_receive},z,t_2\right), \\
&\ y\neq z. 
\end{split}
\end{align}

Apparently, where one lives has nothing to do with winning awards. However, the two properties often appear together in one entity, causing us to calculate it tens of thousands of times. In addition, not only $\mathsf{before}$, we also try to calculate whether these instantiations satisfy all other temporal predicates considered in this paper, resulting in several times the computational effort.

If we can quickly evaluate whether matched subgraphs, i.e., instantiations, can generate potentially valuable constraints, we can speed up the whole statistical process.
Let us consider Bernoulli's law of large numbers:
\begin{align}
    \lim_{n \to \infty}P(|\frac{\mu_n}{n}-p|  \leq \varepsilon) = 1,
\label{f:large law}
\end{align}
where $\mu_n$ is the number of occurrences of event A in $n$-fold Bernoulli experiments, $p$ is the probability of event A occurring in each experiment, and $\varepsilon$ is any positive number.
The frequency $\mu_n$ of event A’s occurrence converges to the probability $p$ of A’s occurrence. That is to say, when we observe enough instantiations of a possible constraint, we can roughly estimate the final confidence score of the constraint by the current confidence.

We now show specific practices for pruning during mining.

\subsection{Two-stage Pruning}
\label{sec:pruning}
Our mining algorithm can be summarized as subgraph matching and confidence calculation. More specifically, we first match subgraphs according to graph patterns. Then for each subgraph we generate all possible constraints, i.e. combining the subgraph with temporal predicates. The process of calculating whether it is a support is to calculate the logical value of the temporal predicate. Our two-stage pruning is performed at the confidence calculation and subgraph matching stages, respectively.

\subsubsection{Pruning at the Confidence Calculation Stage}
As already explained above, in the confidence calculation stage we first concatenate the matched subgraphs with all temporal predicates to generate all potential constraints. Then, we compute the confidence for each potential constraint.
When the mining process begins for a while, we have obtained incomplete statistics of support and instantiations for many possible constraints. We consider that the constraints whose instantiations $n$ exceed $\alpha$ times the support threshold $\theta_{freq}$ has occurred enough times. Applying the law of large numbers Eq.~(\ref{f:large law}), the final confidence $p$ is approximately $\frac{\mu_n}{n}$, that is, the confidence calculated from incomplete statistics.
If $\frac{\mu_n}{n}$ is less than $\beta$ times the candidate confidence threshold $\theta_{c2}$, which is a lower value compared to the final confidence threshold $\theta_{c1}$, we consider this possible constraint to be a bad constraint. Here $\alpha$ and $\beta$ are adjustable hyperparameters. Once a possible constraint is judged to be bad, we no longer calculate metrics for it.

Take Constraint~\ref{worthlessconstraint} as an example to illustrate how the pruning at this stage is carried out. After we have seen more than $\alpha \times \theta_{freq}$ subgraphs matching Constraint~\ref{worthlessconstraint}, we calculate that the confidence of this constraint is currently much lower than $\beta \times \theta_{c2}$. Then, we no longer calculate whether subgraphs matching Constraint~\ref{worthlessconstraint} satisfy the temporal predicate $\mathsf{before}$. In the same way, we also get that $\mathsf{residence}$ and $\mathsf{award\_receive}$ do not satisfy any other temporal predicate, so the confidence calculation process is also pruned.

\subsubsection{Pruning at the Subgraph Matching Stage}
After a period of pruning in the confidence calculation stage, we have statistically obtained which subgraphs have the potential to generate valuable constraints and which do not. We then use these statistics to guide pruning in the subgraph matching stage. 

The core idea of pruning at this stage is that if we count enough times, more precisely $\gamma$ times the support threshold $\theta_{freq}$, that a property appears in matched subgraphs (instantiations), but all instantiations containing it are considered unlikely to generate valuable constraints by our estimation at the previous stage, then the property is considered bad, such as $\mathsf{music.artist.home\_page}$ in Freebase. Here $\gamma$ is an adjustable hyperparameter. Once a property is judged to be bad, the property (edge) is directly pruned when matching subgraphs.

In summary, our pruning targets are dynamically determined during the mining process. As the statistical information continues to increase, more and more possible constraints and edges can be pruned, so the speed improvements would become more and more significant.

\subsection{Theoretical Analysis of Running Efficiency}
In this subsection, we theoretically analyze the running efficiency improvement that the pruning strategy can bring. The main computational cost comes from the subgraph matching and confidence calculation. The number of operations required for subgraph matching process is $\begin{matrix} \sum_{v}^V d^2(v) \end{matrix}$ with structural pattern (a) and $\begin{matrix} \sum_{v}^V\sum_{n}^{N(v)} d^2(v)\times d(n) \end{matrix}$  with structural pattern (b), respectively. $V$ is the vertex set of KG and $v$ is the vertex in set $V$. $d$ is the degree of the vertex. $N(v)$ is the neighbor vertex set of $v$ and $n$ is the neighbor vertex of $v$ in set $N(v)$. For each matched subgraph, the time interval algebra needs to be calculated when computing the corresponding evaluation indicators. The time interval algebra calculation can be completed in constant time $k$. Therefore, the computational cost of constraint mining is $\begin{matrix} \sum_{v}^V k \times d^2(v) \end{matrix}$ for structural pattern (a) and $\begin{matrix} \sum_{v}^V\sum_{n}^{N(v)} k \times d^2(v)\times d(n) \end{matrix}$  for structural pattern (b), respectively.

For structural pattern (a), the pruning strategy optimizes $k$ and $d^2(v)$ in the computational cost, respectively. The pruning strategy at the confidence calculation stage will reduce $k$. Specifically, for each subgraph, it is originally necessary to calculate whether it satisfies six temporal relationships. After pruning, only 1 or 0 kind of temporal relationship need to be calculated. Pruning at the subgraph matching stage will cause $d^2(v)$ to drop for each vertex. In the most extreme case, if all edges of a vertex are pruned then $d^2(v)$ will drop to $d(v)$. Generally speaking, after pruning the number of operations required for subgraph matching for each vertex is between $d^2(v)$ and $d(v)$. For structural pattern (b), the pruning strategy optimizes $k$ and $d^2(v) \times d(n)$ in the computational cost in the same way. The specific running time will be shown in Section~\ref{sec:PaTecon+vs}.


\section{Experiments}
\label{sec:exp}

We conducted four groups of experiments. 
In the first group of experiments, we compared constraints mined by PaTeCon+ with handwritten constraints, for there is currently no automatic way to mine temporal constraints. We used them for conflict detection on our new benchmarks and compare the performance. In the second group of experiments, we evaluated constraints measured by entity-level confidence and fact-level confidence. In the third group of experiments, we compared PaTeCon+ with PaTeCon to demonstrate the efficiency improvement brought about by our newly proposed pruning algorithm. Finally, in the fourth group of experiments, we expanded the application scenario and used the constraints mined by PaTeCon+ to to enhance Temporal Knowledge Graph Completion (TKGC) models.

Our experimental results show that:
\begin{enumerate}
    \item PaTeCon+ is highly effective in generating valuable temporal constraints on different KGs requiring no human intervention. The constraints mined by PaTeCon+ have also been shown to be valuable in the conflict detection task.
    \item Entity-level confidence can more accurately reflect the regularity between temporal facts than fact-level confidence.
    \item  The pruning strategy implemented in PaTeCon+ allows us to run  on KGs with more than 10M facts and 1 hundred properties in a matter of minutes. While PaTeCon took about hours for them. 
    \item Constraints mined by PaTeCon+ are effective in enhancing TKGC models by filtering out predicted facts that conflict with the original KG..
\end{enumerate}
\label{results}
\subsection{Experimental Setup}
\subsubsection{Benchmarks}
We utilized \citet{AAAI17}'s WD50K dataset from Wikidata to compare PaTeCon+ with it. To make our evaluations more applicable in real-world scenarios and manageable, we created two datasets, WD27M and FB37M, using two different KGs, respectively.

For WD27M, we expanded the list of properties to include additional temporal properties. We gathered all the properties associated with facts having time qualifiers such as $\mathsf{point\_in\_time}$, $\mathsf{start\_time}$, and $\mathsf{end\_time}$, to acquire a set of 93 temporal properties, while 141 non-temporal properties were collected for persons and organizations. We extracted all the facts with the 93+141 properties from the 2019-01-28 Wikidata dumps\footnote{\url{https://archive.org/download/wikibase-wikidatawiki-20190128}} to evaluate our method's performance.

Since Freebase's schema does not explicitly provide a list of time qualifiers, we followed \citet{AAAI17} and utilized 6 temporal properties for FB37M. We collected all the entities described by these properties and gathered all their one-hop facts from the latest Freebase dumps\footnote{\url{https://developers.google.com/freebase}} for evaluation. The statistics for the three datasets are presented in Table~\ref{tab:datasets}, as shown below.

\begin{table}[htb]
    \centering
    \caption{Dataset overview.}
    \begin{tabular*}{\linewidth}{LRRR}
        \hline\hline
        Datasets & WD50K & WD27M & FB37M\\
        \hline
        Entities & 17,176 & 7,224,869 & 10,178,664\\
        Facts & 50,000 & 27,312,354 & 37,939,422\\
        \quad Temporal & 50,000 & 7,471,929 & 2,989,799 \\
        \quad Others & 0 & 19,840,425 & 34,949,623\\
        Properties & 6 & 234 &  2,492\\
        \quad Temporal & 6 & 93 & 6 \\
        \quad Others & 0 & 141 & 2,486\\
        \hline\hline
    \end{tabular*}
    \label{tab:datasets}
\end{table}

Furthermore, compared to previous work~\cite{chen2023PaTeCon}, in this paper, we semi-automatically annotate WD27M and FB37M to obtain two wrong fact sets WD-411 and FB-128. The statistic of these two datasets is shown in Table~\ref{tab:benchmark-1}. With these two annotated datasets, we can finally automatically evaluate the conflict detection task. However, it is very difficult to manually annotate wrong facts in KG. Imagine that for a KG of generally good quality, annotating 100 facts may only get one wrong fact or even none at all.
Below we propose a method for semi-automatically annotating wrong facts.

\begin{table}[htb]
    \centering
    \caption{The statistic of annotated datasets.}
    \begin{tabular*}{\linewidth}{LRR}
    \hline\hline
    Datasets & WD-411 & FB-128 \\
    \hline
    Total Facts & 822   & 256\\
    \quad Wrong Facts & 411   & 128   \\
    \quad Correct Facts & 411   & 128   \\
    Property  & 69    & 6     \\
    \hline\hline
    \end{tabular*}
    \label{tab:benchmark-1}
\end{table}

We have explained above that it is very inefficient for annotators to check facts one by one according to fact sources. If we can quickly filter out which facts are most likely to be wrong, it will greatly speed up the annotating process. The idea of our method is that as the KG is continuously updated, wrong facts may have been corrected. Therefore, as long as the old version of KG is compared with the new version and the facts that are deleted are counted, the facts with high error probability can be obtained. For Wikidata, the 2019-01-28 dumps used in this paper were compared with the latest data on the Wikidata official website. As for Freebase, which has ceased operation, we compared the fact that it can be aligned with facts in Wikidata with the latest data on Wikidata official website. The obtained set of candidate wrong facts was checked by two KG experts against the source of the facts and annotated as correct or wrong. We made the property distribution in the annotated dataset as consistent as possible with the large datasets WD27M and FB37M. The distribution of properties for the two annotated datasets has been shown in Table~\ref{tab:benchmark-2}. Finally, we added facts annotated correctly in a 1:1 ratio based on the number of wrong facts.

\begin{table*}[htb]
    \centering
    \caption{Property distribution of wrong facts in WD-411 and FB-128.}    
    \begin{tabular*}{\linewidth}{LRLR}
    \hline\hline
    \multicolumn{2}{C}{WD-411} & \multicolumn{2}{C}{FB-128}\\
    \cline{1-2} \cline{3-4}
    Property & Facts & Property & Facts\\
    \hline
    P569(date of birth) & 159 & people.person.date\_of\_birth & 46\\
    P570(date of death) & 103 & people.person.spouse\_s & 28\\
    P54(member of sports team) & 55 & sports.pro\_athlete.teams & 22\\
    P39(position held) & 9 & people.person.date\_of\_death & 20\\
    P166(award received) & 8 & sports.sports\_team.coaches & 8\\
    ... & ... & people.person.employment\_history & 4\\
    \hline\hline
    \end{tabular*}
    \label{tab:benchmark-2}
\end{table*}

\subsubsection{Evaluation Metrics and Setting}
To evaluate the quality of the mined constraints and hand-crafted constraints, we design 3 grades as follows: 
\begin{itemize}
\item \textbf{C}: The constraint is \underline{c}orrect.
\item \textbf{M}: The constraint has some \underline{m}erit, but there are also obvious exceptions.
\item \textbf{W}: The constraint is \underline{w}rong. The involved facts are not temporally related or should be restricted by an opposite predicate.
\end{itemize}
In our evaluation, each constraint is scored by 3 experienced annotators separately. In cases where a consensus cannot be reached, indicated by three annotators rating the constraint as \textbf{C}, \textbf{M}, and \textbf{W}, we assign a quality score of \textbf{M}. We evaluate all generated constraints on WD50K, our dedicated dataset, while for practical large-scale datasets, we randomly select 50 constraints for evaluation. We report the number of detected possible conflicts and the quality rates of PaTeCon+. 

To evaluate the effectiveness of conflict detection using mined constraints and hand-crafted constraints, we first detect conflicts on WD27M and FB37M, and calculate the recall on WD-411 and FB-128. In our setup, only conflicting pairs are detected and no resolution is performed, so precision is not calculated, and a wrong fact contained in a conflicting pair is considered a successfully recalled example. In addition, our annotated datasets WD-411 and FB-128 contain correct and wrong facts, which can provide evaluation for future embedding-based methods to resolve conflicts or judge whether facts are correct or wrong.

In order to report the effectiveness of detecting conflicts between new facts and the original KG, we use the same metrics as the selected baseline method TiRGN~\cite{TiRGN}, namely Mean Reciprocal Ranks (MRR) and Hits@1/3/10 (the proportion of correct test cases that are ranked within top 1/3/10).
\subsubsection{Environments and Parameters}
PaTeCon+'s results were achieved using a Python implementation running on a personal workstation with an Intel Xeon E5-1607 v4 @3.10GHz CPU and 128GB RAM, with a single thread. For the threshold values, we simply set $\theta_{freq}=20$ in WD50K and $\theta_{freq}=100$, $\alpha=0.5$, $\beta=0.8$, $\gamma=5$ in WD27M and FB37M, $\theta_{c_1}=0.5$, $\theta_{c_2}=0.9$ in all datasets.

\subsection{Mined Constraints versus Hand-crafted Constraints}
Table~\ref{tab:WD50K} presents a comparison between the hand-crafted constraints in \citet{AAAI17} and the PaTeCon+ mined constraints. It is noteworthy that 25\% of the hand-crafted constraints have quality issues, while all the mined constraints are deemed correct. These results demonstrate that PaTeCon+ outperforms human experts in terms of specifying the granularity of constraints.

\begin{table}[htb]
    \centering
    \caption{Statistics about the quality grades on WD50K.}
    \begin{tabular*}{\linewidth}{LRRRR}
    \hline\hline
        Method & C & M & W & Total\\ \hline
        \citet{AAAI17} & 9 & 3 & 0 &12\\
        PaTeCon+ & 13 & 0 & 0 & 13\\ \hline\hline
    \end{tabular*}

    \label{tab:WD50K}
\end{table}

In the context of WD50k, the analysis of our statistics reveals that \citet{AAAI17}'s constraints classify approximately $10\%$ of all fact combinations as potential conflicts. Specifically, their constraints identify 107,104 possible conflicts, out of which the noisy constraint exemplified in Example~\ref{eg:raw_rule_sports_teams} covers $96.8\%$ ($103, 662$) of them. On the other hand, our method detects 971 conflicts, and 967 of them correspond to the conflicts identified by \citet{AAAI17}. With regard to inter-rater reliability, the Fleiss's $\kappa$ score on WD50K is $0.10$.

To test the effectiveness of conflict detection using mined constraints and  hand-crafted constraints, we first performed conflict detection on WD27M and FB37M, and then leveraged the annotated data in WD-411 and FB-128 to compute the recall of wrong facts in different systems. We implemented a simple conflict detection system by detecting which two facts violate the constraints. Since \citet{AAAI17} did not provide the source code, we ran the source code in their journal paper~\cite{TECOREpvldb17} to do conflict detection. In addition, as their system cannot run on such large-scale KGs, we provided subgraphs for their system on WD27M and FB37M, respectively. These two subgraphs consist of maximal connected subgraphs where each entity resides in WD-411 and FB-128, ensuring that all information used to infer the truth of a fact is contained in the subgraph.

\begin{table}[htb]
    \centering
    \caption{The performance of conflict detection on WD-411 and FB-128. The results are shown with recall (\%).}    
    \begin{tabular*}{\linewidth}{LRR}
    \hline\hline
    Method & WD-411 & FB-128 \\
    \hline
    \citet{AAAI17}     & 23.11  & 8.59   \\
    PaTeCon+   & 51.34  & 19.53 \\
    PaTeCon+ w/ fact-level confidence & 57.91  & 19.53 \\
    \hline\hline
    \end{tabular*}
    \label{tab:recall}
\end{table}

Table~\ref{tab:recall} presents a comparison between hand-crafted constraints and  mined constraints applied on the conflict detection task. The recall of PaTeCon+ with fact-level confidence will be explained in Section~\ref{sec:confidencevs}. It can be seen that the recall of PaTeCon+ is much better than \citet{AAAI17}. This is of course due to the fact that the number of mined constraints is far greater than that of hand-crafted constraints. On the other hand, it also shows that the mined constraints can indeed detect conflicts that contain wrong facts. Overall, these results demonstrate the effectiveness of mining constraints for conflict detection.
\subsection{Entity-level Confidence versus Fact-level Confidence}
\label{sec:confidencevs}
In order to illustrate the difference between the constraints measured by entity-level and fact-level confidence,  we illustrate their differences from two dimensions, the quality of constraints and the performance on  conflict detection.

Table~\ref{tab:quality} illustrates quality grades of the mined constraints measured by entity-level and fact-level confidence. Here, we still randomly select 50 constraints for manual evaluation. In summary, 86\% and 96\% of the mined constraints measured by the entity-level confidence are considered valuable on WD27M and FB37M, respectively, and about half of the constraints (50\% and 58\% on WD27M and FB37M, respectively) of the mined constraints are considered correct and do not need further modifications. In contrast, 90\% and 94\% of the mined constraints measured by fact-level confidence are considered valuable on WD27M and FB37M, which is almost the same as the entity-level confidence. However, the percentages of constraints (20\% and 46\% on WD27M and FB37M, respectively) considered correct drop significantly and more constraints become ambiguous.

\begin{table*}[htb]
    \centering
    \caption{Statistics about quality grades of mined rules with fact-level confidence and entity-level confidence  on WD27M and FB37M.}
    \begin{tabular*}{\linewidth}{LRRRRRR}
    \hline\hline
    \multirow{2}{*}{Method}    & \multicolumn{3}{C}{WD27M} & \multicolumn{3}{C}{FB37M}\\
    \cline{2-4} \cline{5-7}
    & C & M & W  & C & M & W\\
    \hline
    PaTeCon+ w/ entity-level confidence  & 50\% & 36\% & 14\%  & 58\% & 38\% & 4\%\\
    PaTeCon+ w/ fact-level confidence  & 20\% & 70\% & 10\%  & 46\% & 48\% & 6\%\\
    \hline\hline
    \end{tabular*}
    \label{tab:quality}
\end{table*}

With regard to inter-rater reliability, the Fless's $\kappa$ scores of PaTeCon+ with entity-level confidence  are $0.29$, $0.22$ on WD27M and FB37M, and  PaTeCon+ with fact-level confidence scores $0.39$, $0.37$, respectively.

\begin{table*}[htb]
    \centering
    \caption{The number of constraints and detected conflicts using entity-level and fact-level confidence measures. Statistics of our old version PaTeCon are also listed.}
    \begin{tabular*}{\linewidth}{LRRRR}
    \hline\hline
    \multirow{2}{*}{Method}    & \multicolumn{2}{C}{WD27M} & \multicolumn{2}{C}{FB37M}\\
    \cline{2-3} \cline{4-5}
        & constraints & conflicts & constraints & conflicts\\
    \hline
    PaTeCon & 709 & 242,538 & 82 & 21,180\\
    PaTeCon+ w/ entity-level confidence & 637 & 208,998 & 78 & 21,167\\
    PaTeCon+ w/ fact-level confidence & 1,128 & 7,669,966 & 118 & 25,090\\
    \hline\hline
    \end{tabular*}
    \label{tab:confidence}
\end{table*}

We combine the recall of wrong facts, the number of mined constraints, and the number of detected conflicts  to interpret the conflict detection performance. Table~\ref{tab:confidence} shows the number of constraints mined by PaTeCon+ with fact-level and entity-level confidence, as well as the conflicts they identified in the large-scale datasets WD27M and FB37M. The statistic of PaTeCon will be used in Section~\ref{sec:PaTecon+vs}. It can be seen that mined constraints with the fact-level confidence measure are several times that of those with the entity-level confidence measure. Therefore, the former also detects more conflicts than the latter. However, simply comparing the number of constraints and conflicts does not tell much. As is shown in Table~\ref{tab:recall}, compared to PaTeCon+ (with entity-level confidence), PaTeCon+ with fact-level confidence improves recall by about 13\% on WD-411 and exactly the same on FB-128. Nevertheless, this result is obtained with about 1.8x constraints, 38x conflicts on WD27M,  and 1.5x constraints, 1.2x conflicts on FB37M, respectively. This means that fact-level confidence measures lead to more constraints, but with lower quality, which is consistent with our theoretical analysis in Section~\ref{sec:entitylev}. Furthermore, the conflicts detected by constraints with fact-level confidence are less likely to contain wrong facts, especially on WD27M.

\subsection{PaTeCon+ versus PaTeCon}
\label{sec:PaTecon+vs}
In this section, we discuss the speed improvement in PaTeCon+ over the previous version PaTeCon. 
The performance of PaTeCon/PaTeCon+ is impacted by several factors, including the number of entities, properties, classes, and the density of the KG under consideration. Given that real-world KGs are often very sparse, the execution time is significantly lower than the time it would take to enumerate all possible combinations of properties and classes. 

Table~\ref{tab:running_time} lists the cost of running PaTeCon and PaTeCon+ on each benchmark. It can be seen that the running time of PaTeCon on KGs with tens of millions of knowledge is acceptable. Moreover, with the pruning strategy, the running efficiency of PaTeCon+ on the two datasets has been greatly improved. It can be seen from the table that the running time of PaTeCon+ on WD27M and FB37M is only about 70\% and 14\% of PaTeCon, respectively.

\begin{table}[htb]
    \centering
    \caption{Running time (s) of PaTeCon, PaTeCon+ and PaTeCon+ w/o  stage 2 pruning on WD27M and FB37M, respectively.}
    \begin{tabular*}{\linewidth}{LRR}
    \hline\hline
    Method & WD27M & FB37M  \\
    \hline
    PaTeCon         & 2,072 & 10,647 \\
    PaTeCon+        & 1,361 & 1,493 \\
    PaTeCon+ w/o pruning stage 2  & 1,737 & 8,909 \\
    \hline\hline
    \end{tabular*}
    \label{tab:running_time}
\end{table}
    
We also conduct ablation experiments to illustrate the effect of pruning at each stage. As shown in Table~\ref{tab:running_time}, without  stage 2 pruning running time is 27\% longer on WD27M and exaggeratedly longer (14\% to 84\% of PaTeCon's running time) on FB37M. This illustrates the effectiveness of stage 2 pruning for filtering properties that do not have temporal associations, especially on FB37M, which has 2,492 properties. Examples of specific properties that are pruned are as follows: 
\begin{align*}
    &music.artist.home\_page, \\
    &base.catalog.cataloged\_composer.music\_catalog, \\ 
    &music.artist.genre.
\end{align*}
It should be noted that we cannot discard stage 1 pruning alone because stage 2 pruning is based on the pruning results and statistical data of stage 1. From Table~\ref{tab:running_time} we can see that stage 1 pruning also brings a significant speed improvement to our method, saving about 15\% of running time compared to PaTeCon on both KGs. In summary, the pruning strategy at each stage plays an important role in speed improvement.

Furthermore, we discuss the effect of the pruning strategy on the mined constraints. As has been shown in Table~\ref{tab:confidence}, using the pruning strategy does result in fewer constraints (10\% on WD27M and 5\% on FB37M) on the final output. This is understandable since we prune based on the law of large numbers. When we set $n$ in Equation~\ref{f:large law} to be relatively small, misjudgments may occur. By changing the hyperparameters of the pruning strategy, we can strike a balance between speed improvement and constraint quantity penalty.

\subsection{Enhancing Temporal Knowledge Graph Completion Models}
Since the facts in the KG are constantly updated and growing, the purpose of this experiment is to test the ability of PaTeCon+ to enhance the quality of newly added facts. The existing KG has passed the test of time, so the fact quality is high, while the error rate of new facts is relatively high. Facts predicted by TKGC models can be a good source of new facts. TiRGN~\cite{TiRGN} and RE-GCN~\cite{REGCN} are two competitive models on the TKGC task. We use the prediction results of TiRGN and RE-GCN on YAGO~\cite{YAGO3} as the source of new facts.

Since the splits of training, validation and test sets in TiRGN and RE-GCN are completely disjoint in time, that is, time on the training set $<$ the validation set $<$ the test set, we first mix the validation and test sets and then re-partition them. We do this re-partition because we believe that newly added facts should not only be the latest facts in time but also missing past facts will continue to be added. And we only mix the validation set and test set, keeping the training set unchanged retains the characteristics of TiRGN and RE-GCN learning from the history and predicting the future. The experimental results in Table~\ref{tab:tkgc} show that TiRGN's performance (MRR: 97.09) and RE-GCN's performance (MRR: 97.69) on our modified dataset are not much different from the results (MRR: 99.30 and 98.80, respectively) reported in their papers.

The process of TiRGN/RE-GCN+Conflict Detection is to get the property prediction results of TiRGN/RE-GCN for each test case (e.g, (\textsf{Barack Obama, ?, India, 2014-04-25})), that is, a ranked list of candidate properties. Then we fill the properties in the ranked list into the prediction slot to form new facts. If the new fact conflicts with the original KG, we adjust the order of the according predicted property in the ranked list to the end. We recalculate the metrics for the reranked predictions.

The experimental results in Table~\ref{tab:tkgc} shows two things. First, we need to know that constraints may wrongly determine that a correct new fact conflicts with the original KG. Therefore, the experimental results do not decrease, indicating that the mining constraints generally have no misjudged conflicts. Second, Hit@1 and MRR still have a good improvement despite the high scores of metrics, which shows that we have indeed detected that many of the high-ranking prediction facts conflict with the original KG. Overall, we also validate the feasibility of enhancing TKGC models with mined constraints.

\begin{table*}[htb]
    \centering
    \caption{Performance (in percentage) for property prediction task on YAGO.}
    \begin{tabular*}{\linewidth}{LRRRR}
    \hline\hline
    \multirow{2}{*}{Method} & \multicolumn{4}{c}{YAGO}    \\
    \cline{2-5}      
          & MRR   & HIT@1 & HIT@3 & HIT@10 \\
    \hline
    RE-GCN & 97.69 & 96.22 & 98.91 & 99.83 \\
    RE-GCN+Conflict Detection & 97.75 & 96.37 & 98.91 & 99.83 \\
    \hline
    TiRGN & 97.09 & 95.54 & 98.23 & 99.66  \\
    TiRGN+Conflict Detection  & 97.41 & 96.12 & 98.29 & 99.68 \\
    \hline\hline
    \end{tabular*}
    \label{tab:tkgc}
\end{table*}

\subsection{Case Study}
In this part, we present highly representative temporal constraints in order to provide a more intuitive view of our proposed mining method.

In the introduction section, we previously discussed the issue of granularity that may arise from constraints that are engineered by humans. As an illustration, Example~\ref{eg:raw_rule_sports_teams} from \citet{AAAI17} showcases such a constraint. To further examine this, we studied the output of PaTeCon+ on WD27M and found that PaTeCon+ is capable of not only capturing the original temporal constraint but also refining it into more dependable sub-constraints. An example of such a refined temporal constraint can be seen in Example~\ref{eg:case0}. 

\begin{example}
\label{eg:case0}
The graph representation of a refined version of ``\textit{One cannot be a member of two sports teams at the same time}'', mined from WD27M. 
\begin{figure}[htb]
    \centering
    \includegraphics[scale=0.7]{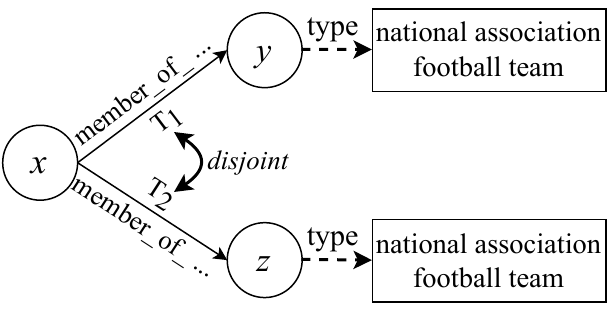}
    \caption{Graph representation of ``\textit{One cannot be a member of two national association football teams at the same time}''.}
    \label{fig:case1}
\end{figure}
\end{example}

Overall, our investigation resulted in the generation of 8 high-quality temporal constraints. These constraints include class restrictions on $y$ and $z$ such as \textsf{national sports team}, \textsf{UCI Continental Team}, among others. This case highlights the necessity of breaking down coarse temporal constraints, which can significantly enhance the quality of constraints.

It is worth noting that PaTeCon+ did not dig out such a refined constraint ``\textit{one cannot play for two association football clubs at the same time}". In other words, y and z cannot be of class \textsf{association football club}. This is mainly due to the fact that loans of players between association football clubs are very common. For example, Figure~\ref{fig:beckham} shows that David Beckham played for LA Galaxy from 2007 to 2012, and was loaned to A.C. Milan from January 2009 to July 2009. If this constraint is used for conflict detection, many fact pairs without conflicts will be found. It can be seen that the constraints mined using statistical learning-based methods can better understand the knowledge representation of KGs than the handwritten constraints.

\section{Related Work}
\label{sec:work}

\subsection{Acquisition of Temporal Knowledge}
Temporal knowledge acquisition has been an active research topic in recent years. Temporal/event KGs such as GDELT~\cite{gdelt}, ICEWS~\cite{icews}, and Event-KG~\cite{EventKG}, along with large-scale general KGs such as Freebase and Wikidata, contain numerous temporal facts. Lots of studies focus on reasoning missing temporal facts via deep learning models ~\cite{Jiangcoling16,KNOWicml17,TTRANSEwww18,RENETemnlp20,LFHaaai21,ding2021automatic,CENacl22,ding2023time,chen2024timeline}. Several of these models have incorporated temporal constraints to improve their reasoning ability. For instance, \citet{Jiangcoling16} adds temporal consistency constraints to their KG completion model, while TIMEPLEX~\cite{TIMEPLEXemnlp20} improves their base model by including additional (soft) temporal constraints such as relation recurrence, ordering, and time gaps. 

Overall, deep learning has led to significant progress in temporal knowledge acquisition, but it also demands the efficient design of explicit temporal constraints.

\subsection{Temporal Conflict Detection and Resolution}
Temporal conflict detection and resolution have only been addressed in a few studies~\cite{RTCbtw11,AKBC14,AAAI17,ETCwise18}. \citet{RTCbtw11} expresses temporal constraints using first-order logic Horn formulas with temporal predicates and employs a scheduling algorithm to resolve conflicts. \citet{AKBC14} debugs temporal KGs using a Markov Logic Network to compute the Maximum a posteriori (MAP) state. \citet{AAAI17} proposes a Markov Logic Network (MLN) approach for reasoning over temporal KGs, but their method mainly relies on hand-crafted temporal constraints with a few AMIE-mined rules. However, AMIE, a general mining approach for RDF KGs, cannot handle quadruple temporal facts with time interval restrictions. In contrast, ETC~\cite{ETCwise18} manually constructs a constraint graph to detect conflicting facts and model the truth inference problem as a maximum weight clique problem.

Despite their efforts, all of the above studies rely heavily on manual constraint engineering and assume that the sampled temporal facts from KGs are correct. They also use specially constructed datasets by adding incorrect facts and assigning confidence to each fact, instead of detecting conflicts from practical KGs.

\subsection{Rule Learning}
Learning rules over a KG is considered a form of statistical relational learning, as noted by \citet{statisticalrelationallearning}. Typically, traditional symbolic-based rule learning methods rely on effective search strategies and pre-defined static evaluation indicators. AMIE~\cite{AMIEwww13} designs three operations: dangling atom, instantiated atom, and closing atom. These operations add various types of atoms to incomplete `rules and rely on pre-defined evaluation metrics (PCA confidence) to remove inaccurate rules. AMIE+~\cite{AMIE+} improves upon this process by revising the rule extending procedure and enhancing the evaluation method used by AMIE. Anyburl~\cite{AnyburlIJCAI19} proposes a framework for efficiently mining rules. It replaces certain entities with variables based on randomly sampled paths to obtain rules. However, Anyburl sticks to standard confidence and just makes minor modifications to counting the number of body atoms. Due to the large size of KGs, all the methods above inevitably face a considerable search space. Therefore pruning techniques are also necessary. Our work is specially designed on search strategy, pruning strategy, and static evaluation metrics,  respectively.

Most rule learning methods are designed for positive rule discovery and ignore negative rules, which can also be called constraints. 
Some studies have focused on identifying constraints in relational data, as demonstrated in \citet{Chupvldb13}. However, the schema-less nature of RDF data and the Open World Assumption (OWA) make it impossible to apply these techniques to KGs. Rudik~\cite{RudikICDE18} provides a solution that can extract both positive and negative rules. The former can be utilized to deduce fresh facts in the KG, while the latter can be beneficial for identifying erroneous triples.

However, these approaches are designed for static KGs rather than TKGs. Very little work~\cite{OmranW019,LCGE} mine the temporal rules on KGs but they only focus on positive rules and not on constraints. AETAS~\cite{Aetas} mines temporal rules to clean web data but relational database techniques can not be applied on RDF KGs as explained in the previous paragraph. Moreover, AETAS focuses on detecting anomalies in temporal values, which is only a subset of temporal constraints. 

\section{Conclusion}
\label{sec:con}
The main focus of this paper is on automating the mining of temporal constraints, where the temporal constraints are the logical assertions of subgraphs in KG on the time dimension. To achieve this, we propose PaTeCon, an approach to generate temporal constraints based on a statistical analysis of pre-defined patterns. Our approach uses two structural patterns that contain 5 temporal predicates and a mutual exclusion predicate represented by $\mathsf{false}$. 

We extended PaTeCon to PaTeCon+ by a two-stage pruning strategy. As our extensive experiments have shown, PaTeCon+ shortened the original operation time to tens of minutes on the KG of tens of millions of facts. In addition, we also present our results on two new benchmarks, demonstrating the effectiveness of our pattern-based constraint mining approach. Finally, we experimented with the newly added fact scenario, which also validated the effectiveness of our mined constraints.

In practical scenarios, PaTeCon+ can be extended to accommodate more intricate requirements. Specifically, the predicates can be broadened to encompass quantitative relationships, such as $t_2 - t_1 \le 10 years$ (where $t_1$ denotes a time point at least ten years before $t_2$). Future work includes exploring extensions to PaTeCon+, such as the expansion of predicates and the amalgamation of extracted constraints. Additionally, statistical pattern-based mining techniques may not fully capture the nuanced meaning of conceptual terms in KGs. Therefore, it is worthwhile to investigate the combination of PaTeCon+ with human experts to obtain constraints that are both efficient and accurate for real-world applications.




\printcredits

\section*{Declaration of competing interest}
The authors declare that they have no known competing financial interests or personal relationships that could have appeared to influence the work reported in this paper.

\section*{Acknowledgements}
This work was supported by the National Natural Science Foundation of China under Grant No. 62272219. The authors would like to thank all the participants of this work and anonymous reviewers.

\bibliographystyle{cas-model2-names}

\bibliography{cas-refs}



\end{document}